\documentclass[runningheads]{llncs}
\usepackage[T1]{fontenc}
\usepackage{graphicx}

\usepackage{hyperref}
\usepackage{amsmath}
\usepackage{makecell}
\usepackage{amssymb}
\usepackage{amsmath}

\usepackage{multirow}
\usepackage{pifont}
\usepackage{booktabs}
\usepackage{tabularx}
\usepackage{array}
\usepackage[misc]{ifsym}

\newcolumntype{C}{>{\hspace{0.5em}}c<{\hspace{0.5em}}}

\usepackage{xcolor}
\definecolor{blue1}{RGB}{220,234,247}
\definecolor{green1}{RGB}{217,242,208}
\definecolor{yellow1}{RGB}{255,255,204}
\usepackage{layout}
\usepackage{adjustbox}
\newcommand{\marktext}[2]{\adjustbox{bgcolor=#1}{\strut #2}}

\begin{document}
\title{Integrating Clinical Knowledge into Concept Bottleneck Models}
\titlerunning{Integrating Clinical Knowledge into CBMs}
\author{Winnie Pang \and  
Xueyi Ke \and
Satoshi Tsutsui \and
Bihan Wen\thanks{Bihan Wen is the corresponding author.} } 
%
%

%
\authorrunning{W. Pang et al.}
\institute{Nanyang Technological University, Singapore \\
\email{winnie001@e.ntu.edu.sg, bihan.wen@ntu.edu.sg} }
\maketitle              %
\begin{abstract}

Concept bottleneck models (CBMs), which predict human-interpretable concepts (e.g., nucleus shapes in cell images) before predicting the final output (e.g., cell type), provide insights into the decision-making processes of the model. However, training CBMs solely in a data-driven manner can introduce undesirable biases, which may compromise prediction performance, especially when the trained models are evaluated on out-of-domain images (e.g., those acquired using different devices). To mitigate this challenge, we propose integrating clinical knowledge to refine CBMs, better aligning them with clinicians' decision-making processes. Specifically, we guide the model to prioritize the concepts that clinicians also prioritize. We validate our approach on two datasets of medical images: white blood cell and skin images. Empirical validation demonstrates that incorporating medical guidance enhances the model's classification performance on unseen datasets with varying preparation methods, thereby increasing its real-world applicability. Our code is available at \url{https://github.com/PangWinnie0219/align_concept_cbm}.

\end{abstract}

\section{Introduction}
The integration of deep learning in medical image analysis has greatly improved prediction performance~\cite{han2017breast,ardila2019end,ran2021deep}. However, the inherent ``black-box'' nature of these deep models often presents challenges in understanding the decision-making processes, thus limiting their acceptance in clinical settings.
The Concept Bottleneck Model (CBM) \cite{koh2020concept} was proposed to address this issue by establishing causal relationships between interpretable concepts and final class predictions, thereby enhancing the explainability of the decision-making process.
For example, in the classification of white blood cells (WBC), CBM facilitates the interpretation of the model's decisions in terms of morphological concepts such as nucleus shape and granule color. Furthermore, CBM has been applied to other medical tasks~\cite{marcinkevivcs2024interpretable,amgad2024population,yan2023towards}.

However, models trained solely on data often inadvertently capture biases present in the training data, leading to diminished performance, especially under domain shifts from training data, such as changes in staining conditions or microscopic devices~\cite{tsutsui2023benchmarking}. 
In contrast to deep learning models, clinical experts can generally achieve reliable classification performance because they are equipped with domain knowledge. When classifying WBC types based on morphological concepts, for example, experts focus on concepts that differentiate between different cell types~\cite{al2018classification,standring2005gray}. While all concepts are clinically relevant to WBCs, certain ones hold more relevance to specific cell types. Experts concentrate on these critical concepts to make decisions. This knowledge helps human experts perform better in recognizing WBCs. Motivated by the above observation, we hypothesize that CBMs can similarly benefit from domain knowledge if appropriately integrated. Then, \textit{how to effectively incorporate domain knowledge into CBMs?} To address this, we present an approach that encourages CBMs to mirror the decision-making process of clinicians based on clinical concepts.

\begin{figure}[tb!]
  \centering
  \includegraphics[trim=0 420pt 0 0, clip, width=\textwidth]{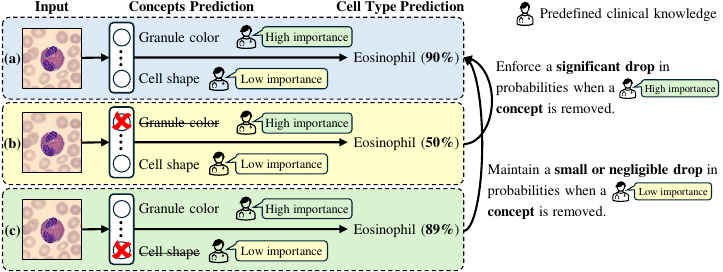}
    \caption{%
    We propose a method to guide concept bottleneck models (CBMs) using knowledge aligned with clinicians' perspectives.
    \protect\marktext{blue1}{\textbf{(a)}:} CBMs predict interpretable concepts (e.g., granule color, cell shape, etc.) and then make a final prediction (e.g., eosinophil) based on them. During training, models usually do not consider the clinical importance of the concepts. Therefore, granule color and cell shape are treated equally despite granule color being a much more important factor for predicting eosinophils.
    \protect\marktext{yellow1}{\textbf{(b)}:} To incorporate clinical knowledge, we enforce the CBM to exhibit a significant drop in cell type prediction probabilities when a clinically important concept is removed from the prediction. For instance, the predicted eosinophil probability should be lower when granule color, a key factor in recognizing eosinophil, is missing.
    \protect\marktext{green1}{\textbf{(c)}:} Conversely, the cell type prediction probabilities should experience a negligible drop when a less clinically important concept is removed from the prediction. For instance, the eosinophil probability should not be affected much when cell shape, which is irrelevant to recognizing eosinophil, is missing.
    }%
  \label{fig:overview}
\end{figure}

In this paper, \textbf{we propose a novel approach to guide the CBM to prioritize the concepts utilized by the clinical experts in decision-making}, as illustrated in Fig.~\ref{fig:overview}. We first identify the importance of each concept utilized by clinical experts in cell type classification. Subsequently, we utilize a perturbation method to determine the importance of each concept from the model in predicting the cell type. We define a concept as \textit{important} if its removal from the prediction results in a substantial variance in prediction probabilities. This approach allows us to determine the importance of all concepts in the model prediction process. Armed with this knowledge, we align the model's perception of concept importance with that of the clinical experts. This is achieved by enforcing a significant drop in prediction probabilities when a clinically important concept is removed from the prediction, thereby ensuring that the model makes its predictions based on concepts aligned with expert prioritization. We conduct experiments on two sets of datasets containing different medical images—WBC images and skin images—and demonstrate that the proposed method can enhance the performance of CBM through the integration of clinical knowledge.

\subsection{Related Work}
Although there are other CBM variants introduced to enhance classification performance or reduce annotation effort~\cite{yuksekgonul2023posthoc,oikarinenlabel,EspinosaZarlenga2022cem}, to the best of our knowledge, no existing work incorporates clinical knowledge into CBMs. However, prior work has explored integrating clinical knowledge into black-box models. For instance, Yin et al.~\cite{yin2021focusing} proposed a regularization module to guide attention maps towards clinically interpretable features like nuclei and fat droplets in histological pattern analysis. Similarly, Zhou et al.~\cite{zhou2023novel} designed a selection module that mimics dermatologists' focus on lesion features such as plaque and scale from noisy backgrounds. In addition, Manh et al.~\cite{manh2022multi} introduced a multi-attribute attention network to guide the model in learning clinically relevant concepts, such as calcification and nodule shape, for predicting thyroid nodule malignancy. 

While the utilization of clinical concepts itself is not a new idea, our novelty lies in the following two aspects. Firstly, while prior work utilize clinical knowledge to guide black-box models, our work focuses on interpretability and thus devises techniques to integrate clinical knowledge specifically into CBMs. Secondly, while prior work defines clinical knowledge in a class-agnostic manner (e.g., nuclei should always be prioritized for any classes), our method enables the definition of more fine-grained knowledge per class. For instance, our method can assign priority to granule color when predicting eosinophils, while emphasizing cytoplasm texture and nucleus shape when predicting monocytes, considering the absence of granules in these cells. Based on these two key differences (focus on CBMs and fine-grained knowledge representations), prior work based on attention~\cite{yin2021focusing,zhou2023novel,manh2022multi} cannot be used as our baselines without substantial modifications because adding these attention mechanisms to CBMs for utilizing fine-grained knowledge is not trivial.

\section{Method}

\begin{figure}[tb!]
  \centering
  \includegraphics[trim=0 396pt 0 0, clip, width=\textwidth]{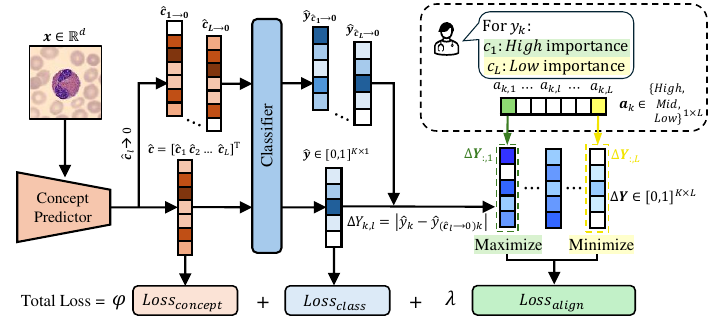}
    \caption{
Integration of Clinical Knowledge into CBM: The lower path illustrates the original CBM, where the final class prediction ($\hat{\mathbf{y}}$)  is based on the prediction of concepts ($\hat{\mathbf{c}}$). The concept importance learned from the model ($\Delta Y$)  is obtained through computing the absolute difference between the prediction probabilities when all concepts are used for prediction ($\hat{\mathbf{y}}$) and when the concept is removed from prediction ($\hat{y}_{\hat{c}_{l}\to0}$). We align the $\Delta Y$  with the concept importance from the clinician's perspective through $Loss_{align}$.  Specifically, we enforce the model to maximize the $\Delta Y_{:,l}$ for the concept that is considered as \textit{High} important by the clinicians, while we constrain the model to minimize the $\Delta Y_{:,l}$ for the concept that is considered \textit{Low} important by the clinicians.}
  \label{fig:architecture}
\end{figure}

Our work is built upon the CBM~\cite{koh2020concept}, which is designed to make final class predictions based on human-interpretable concepts. Formally, the model is abstracted as $x \mapsto \hat{c} \mapsto \hat{y}$, trained on 
$n$ data points of $\{( \text{image } x^{(i)}, \text{concepts } c^{(i)}, \text{target } y^{(i)})\}_{i=1}^{n}$,
where $\hat{c}$ and $\hat{y}$ denote the predicted $c$ and $y$, respectively. One-hot vector $y \in \{0,1\}^{K \times 1}$ is used to represent $K$ classes. Moreover, to represent $L$ concepts, we formulate $c$ as a concatenation of probability vectors, $c = \begin{bmatrix} c_1 & c_2 & \cdots & c_L \end{bmatrix}^\top$ with corresponding prediction $\hat{c} = \begin{bmatrix} \hat{c}_1 & \hat{c}_2 & \cdots & \hat{c}_L \end{bmatrix}^\top$. The $l$-th concept $c_l \in \{0,1\}^{N_l}$ represents the ground truth value out of $N_l$ possible values. For instance, the concept of \texttt{cytoplasm-color} is the 5th concept and has the three possible values \texttt{\{light-blue, purple-blue, blue\}}, where its ground truth of \texttt{blue} is represented as $c_5 = [0,0,1]^\top$, while its prediction can be $\hat{c}_5 = [0.1,0.1,0.8]^\top$. When we \textit{remove} this concept from the input of the classifier ($\hat{c}_{5}\to 0$), we fill zeros like $[0,0,0]^\top$, and denote the resulting class prediction after the removal as $\hat{y}_{\hat{c}_{5}\to0}$. The architecture of the proposed method is illustrated in Fig.~\ref{fig:architecture}.

\subsection{Align Model's Concept Importance with Clinical Knowledge}

\textbf{Clinical Knowledge Representation}: We rank $L$ concepts based on a three-tier scale of importance: \textit{High}, \textit{Mid}, and \textit{Low}, reflecting domain knowledge. Specifically, we assign a concept importance vector $\alpha_k \in {\{\textit{High}, \textit{Mid}, \textit{Low}\}}^{1\times L}$ per final prediction class $k \in \{1,...,K\}$. For example, pathologists often identify eosinophils based on the red color of granules~\cite{standring2005gray}. In this case, the concept \texttt{granule-color} is marked as \textit{High} for the class of eosinophils. If an eosinophil is assigned $k=2$ and \texttt{granule-color} is indexed at $l=7$, then $\alpha_{2,7}=\textit{High}$.

\textbf{Alignment}: 
Our aim is to align the importance of knowledge-based concepts, as specified above, with the importance perceived by the model. To quantify the model's perception, we calculate the impact of each concept on the final class prediction using a perturbation method. Specifically, we first compute the class probabilities when $l$-th concept is removed from the prediction $\hat{y}_{\hat{c}_{l}\to0}$, for $l=1,..,L$. Then, we calculate the absolute difference between $\hat{y}$ (the original predicted probability before removing a concept) and $\hat{y}_{\hat{c}_{l}\to0}$, denoted as:
\[
\Delta Y_{k,l} := \left| \hat{y}_{k} - \hat{y}_{({\hat{c}_{l}\to0})k}
 \right| \in [0,1]
\]
where $\hat{y}_{k}$ is the predicted probability for class $k$ when all concepts are used for prediction, and $\hat{y}_{({\hat{c}_{l}\to0})k}$ is the predicted probability for class $k$ when $l$-th concept is removed from the input to the class prediction layer. 

We consider concepts whose removal significantly alters the class prediction probability as important according to the model. Our objective is to align these important concepts with those ranked as \textit{High} by clinical experts. To accomplish this, for concepts ranked as \textit{High} importance by experts, each denoted by $l_H$, we employ the L1 loss to maximize the $\Delta Y_{{:},l_H} \in [0,1]^{K \times 1} $: 

\[
\mathcal{L}_{\text{Align(High)}} = \sum_{l_H} \sum_{k} \left| 1 - \Delta Y_{k,l_H} \right|
\]

Conversely, denoting a \textit{Low} importance concept as $l_L$,
we seek to minimize  $\Delta Y_{{:},l_L}$ for the concepts ranked as  \textit{Low} importance by experts:
\[
\mathcal{L}_{\text{Align(Low)}} = \sum_{l_L} \sum_{k} \left| \Delta Y_{k,l_L} \right|
\]

Our loss function is then formulated as:
\[
\mathcal{L} = \varphi\mathcal{L}_c + \mathcal{L}_y + \lambda \left(\mathcal{L}_{Align(Low)} + \mathcal{L}_{Align(High)} \right)
\]
where $\mathcal{L}_c$ and $\mathcal{L}_y$ represent the cross-entropy loss functions for concept and class prediction, respectively. The hyperparameters $\varphi$ and $\lambda$ control the weights assigned to $\mathcal{L}_c$ and $\mathcal{L}_{Align}$, respectively.

\section{Experiments}
To evaluate the effectiveness of our approach to integrate clinical knowledge into CBMs, we assess its performance on two types of medical image datasets: WBC and skin. We evaluate the classification performance on both in-domain and out-of-domain datasets.

\begin{figure}[tb!]
  \centering
  \includegraphics[trim=0 432pt 0 0, clip, width=1\textwidth]{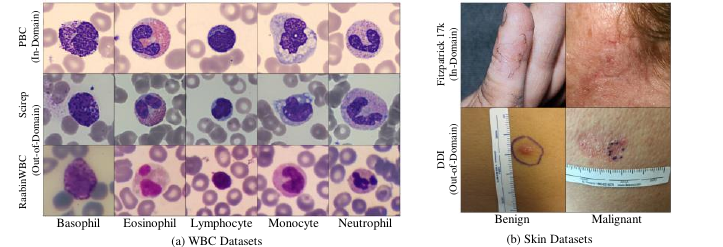}
    \caption{(a) Examples for each cell type from the in-domain dataset (PBC), as well as the out-of-domain datasets (Scirep and RaabinWBC). (b) Skin images for \textit{Benign} and \textit{Malignant} classes from Fitzpatrick 17k (in-domain) and DDI (out-of-domain) datasets.}
  \label{fig:datasets}
\end{figure}

\textbf{(1) WBC datasets}: We used the PBC~\cite{acevedo2020dataset} dataset with concept annotations from the WBCAtt dataset~\cite{tsutsui2023wbcatt}, following their train/val/test splits~\cite{tsutsui2023wbcatt}. Subsequently, we trained the concept predictor ($L=11$) and classifier ($K=5$) for WBC classification. Performance evaluation was conducted on the PBC dataset's testing set for in-domain data and on two other out-of-domain datasets: Scirep~\cite{li2023deep}($n=2019$) and RaabinWBC~\cite{kouzehkanan2022large} ($n=4339$). These out-of-domain datasets contain WBC images under various staining conditions and using different imaging devices. Fig.~\ref{fig:datasets} (a) displays example images. \textbf{Collection of clinical knowledge:} We consulted a pathologist to rank the importance of each concept from her perspective when classifying the WBC images. The pathologist was given 50 images (10 images per WBC class) to rank the importance level on a three-tier scale: \textit{High}, \textit{Mid}, \textit{Low}. We averaged the importance level of each concept for each WBC class, which is used as our clinical knowledge, as shown in Supp. Fig~\ref{fig:delta_y_heatmap}(a).

\textbf{(2) Skin datasets}: 
We train a skin disease classification model ($K=2$) using images from the Fitzpatrick 17k dataset~\cite{groh2021evaluating} ($n=910$) with concept annotations sourced from the SkinCon dataset~\cite{daneshjou2022skincon}. Following the methodology outlined in~\cite{daneshjou2022skincon}, we include only the 22 concepts with at least 50 images ($L=22$). Subsequently, we evaluate the model's performance on two datasets: the Fitzpatrick 17k images without concept annotations serve as the in-domain dataset ($n=3479$), and the Diverse Dermatology Images (DDI) dataset~\cite{daneshjou2022disparities} ($n=656$) serves as the out-of-domain dataset. Since DDI images contain \textit{Benign} and \textit{Malignant} classes only, we utilize the Fitzpatrick 17k images with these two classes for training and testing. Example of images from these datasets are shown in Fig.~\ref{fig:datasets}(b).
\textbf{Collection of clinical knowledge:} We extracted the clinical knowledge (available within the code) from the discussion in the original paper~\cite{daneshjou2022skincon}.

\subsection{Results}

\textbf{Evaluation Metrics}: Due to the class imbalance in the datasets, we use the macro F1 score, which calculates the harmonic mean of precision and recall, instead of plain accuracy.

\begin{table}[tb!]
\centering
\caption{F1 score (mean $\pm$ 95\% CI) of the WBC type classification on cross-dataset testing, with (\textit{Ours}) and without (\textit{Baseline}) the integration of clinical knowledge across various backbone architectures. Generally, incorporating clinical knowledge with the model can enhance performance on out-of-domain dataset evaluations. VGG16 and ViT-B/16 are the backbones of concept predictors, while Linear, MLP(20), MLP(128) refers to the type of classifier with their details available in Supp. Table~\ref{tab:linear}.}
\label{tab:main_table}
\begin{adjustbox}{width=\textwidth}
\begin{tabular}{ccccccc}
\toprule
\multirow{3}{*}{\textbf{\begin{tabular}[c]{@{}c@{}}CBM\\Backbone+Classifier  \end{tabular}}} & \multicolumn{2}{c}{\textbf{In-domain Dataset}} & \multicolumn{4}{c}{\textbf{Out-of-domain Datasets}}                           \\
\cmidrule(lr){2-3} \cmidrule(lr){4-7}
                                                                              & \multicolumn{2}{c}{\textbf{PBC}}               & \multicolumn{2}{c}{\textbf{Scirep}}  & \multicolumn{2}{c}{\textbf{RaabinWBC}} \\
\cmidrule(lr){2-3} \cmidrule(lr){4-5} \cmidrule(lr){6-7}
                                                                              & Baseline         & Ours       & Baseline  & Ours    & Baseline   & Ours       \\
\midrule
VGG16+Linear                                                                     & $99.68_{\textsubscript{$\pm$0.06}}$             & $99.63_{\textsubscript{$\pm$0.04}}$            & $68.99_{\textsubscript{$\pm$3.35}}$     & $\mathbf{70.22}_{\textsubscript{$\pm$0.28}}$ & $51.01_{\textsubscript{$\pm$2.40}}$      & $\mathbf{54.45}_{\textsubscript{$\pm$5.59}}$  \\
VGG16+MLP(20)                                                                    & $99.69_{\textsubscript{$\pm$0.04}}$             & $99.73_{\textsubscript{$\pm$0.02}}$            & $66.91_{\textsubscript{$\pm$3.95}}$     & $\mathbf{70.02}_{\textsubscript{$\pm$6.48}}$ & $52.06_{\textsubscript{$\pm$5.83}}$      & $\mathbf{53.29}_{\textsubscript{$\pm$4.77}}$  \\
VGG16+MLP(128)                                                                    & $99.78_{\textsubscript{$\pm$0.04}}$             & $99.67_{\textsubscript{$\pm$0.06}}$            & $66.31_{\textsubscript{$\pm$3.73}}$     & $\mathbf{70.50}_{\textsubscript{$\pm$2.94}}$ & $54.45_{\textsubscript{$\pm$3.14}}$      & $\mathbf{58.40}_{\textsubscript{$\pm$5.55}}$  \\
\midrule
ViT-B/16+Linear                                                                  & $99.61_{\textsubscript{$\pm$0.12}}$             & $99.69_{\textsubscript{$\pm$0.07}}$            & $75.24_{\textsubscript{$\pm$5.88}}$     & $\mathbf{77.04}_{\textsubscript{$\pm$1.56}}$ & $41.19_{\textsubscript{$\pm$10.68}}$    & $\mathbf{47.27}_{\textsubscript{$\pm$6.46}}$  \\
ViT-B/16+MLP(20)                                                                 & $99.42_{\textsubscript{$\pm$0.41}}$             & $99.67_{\textsubscript{$\pm$0.06}}$            & $71.54_{\textsubscript{$\pm$2.22}}$     & $\mathbf{75.45}_{\textsubscript{$\pm$1.75}}$ & $40.03_{\textsubscript{$\pm$9.05}}$      & $\mathbf{48.13}_{\textsubscript{$\pm$5.17}}$  \\
ViT-B/16+MLP(128)                                                                 & $99.72_{\textsubscript{$\pm$0.02}}$             & $99.63_{\textsubscript{$\pm$0.04}}$            & $76.39_{\textsubscript{$\pm$4.54}}$     & $\mathbf{80.15}_{\textsubscript{$\pm$3.19}}$ & $43.64_{\textsubscript{$\pm$6.67}}$      & $\mathbf{49.75}_{\textsubscript{$\pm$2.39}}$  \\
\bottomrule
\end{tabular}
\end{adjustbox}
\end{table}

\begin{figure}[h!]
  \centering
  \includegraphics[width=0.9\textwidth]{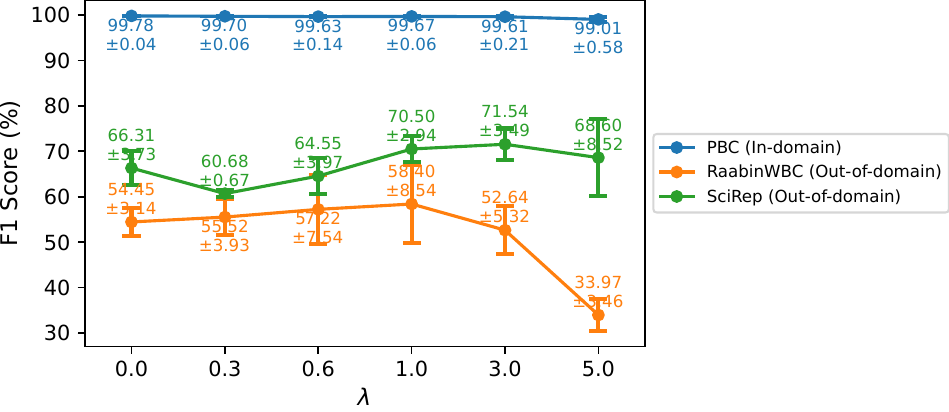}
    \caption{The effect of $\lambda$ in loss function for WBC classification, using VGG + MLP(128).}
  \label{fig:lambdas}
\end{figure}

\begin{figure}[h!]
  \centering
  \includegraphics[trim=0 309pt 0 0, clip, width=0.9\textwidth]{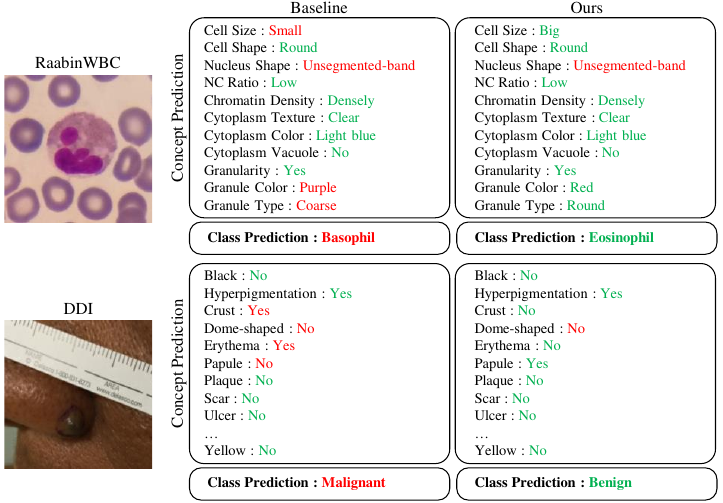}
    \caption{Qualitative results on the out-of-domain WBC and skin datasets demonstrate that our method improves concept predictions, leading to correct class predictions.}
  \label{fig:results}
\end{figure}

\begin{figure}[h!]
  \centering
  \includegraphics[width=\textwidth]{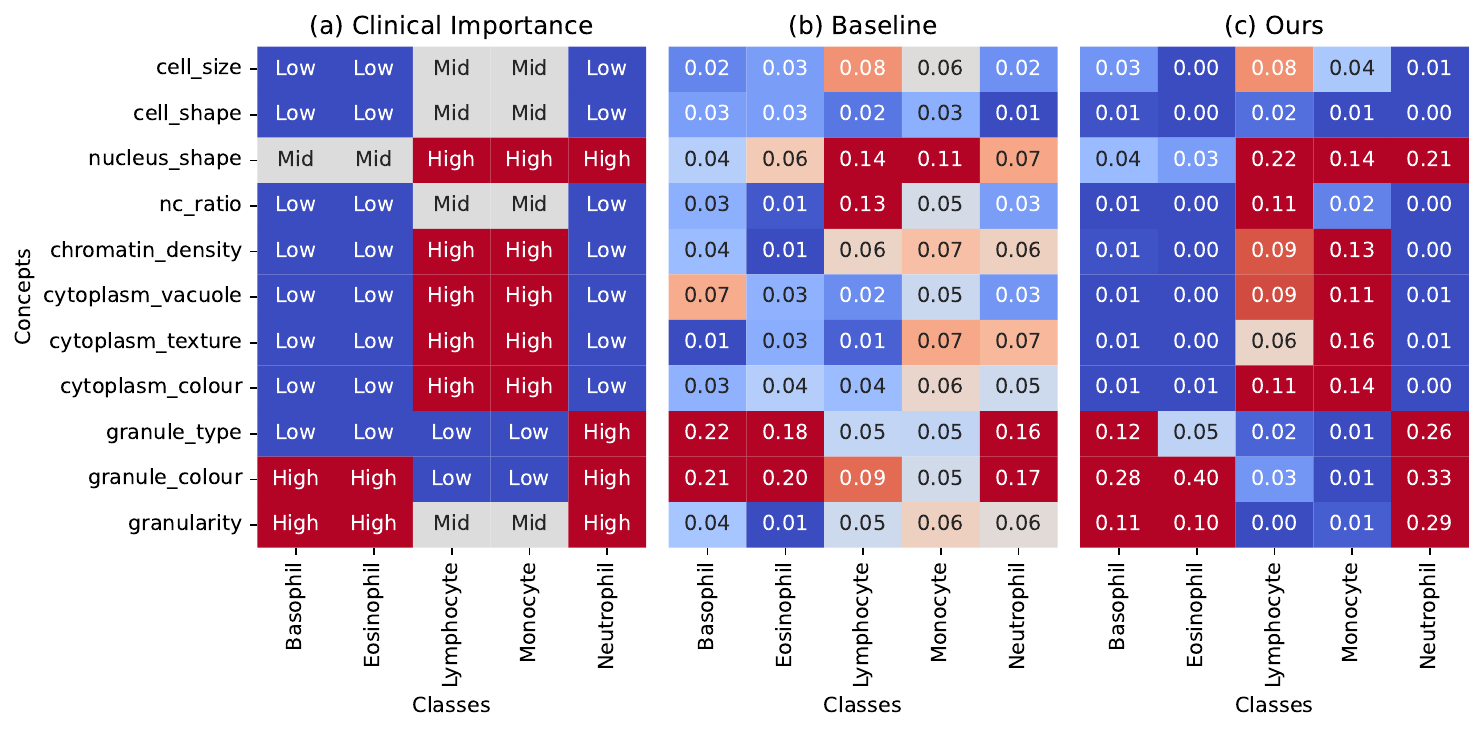}

    \caption{(a) The importance level of the concepts focused on by a pathologist in classifying WBC images into their types. (b), (c) Concept importance learned by the models (Baseline vs. Ours) for WBC classification. The concept importance learned by (c) our model is better aligned with the clinical importance.}
  \label{fig:delta_y_heatmap}
\end{figure}

\textbf{WBC Classification}: We compared the performance of the CBM with (\textit{Ours}) and without (\textit{Baseline}) the integration of clinical knowledge across various backbone architectures. \footnote{\label{note1}Refer to Supp. Table~\ref{tab:linear} and Supp. Table~\ref{tab:implementations} for more details on the classifiers architectures and the implementation details, respectively.} As shown in Table~\ref{tab:main_table}, on the in-domain dataset, the CBM consistently performs well across all backbones, achieving an average F1-score exceeding 99\%. However, when evaluated on the out-of-domain datasets, a noticeable decline in performance is observed, especially on RaabinWBC. 
The decline suggests that the model's performance is significantly affected when tested on images prepared with different staining conditions and imaging devices (Fig.~\ref{fig:datasets}(a)). Notably, integrating clinical knowledge into the CBM results in an improvement in out-of-domain classification performance across all employed backbones. This improvement underscores the efficacy of guiding the model to base its predictions on concepts prioritized by domain experts, thereby enhancing its adaptability and robustness in real-world scenarios.The qualitative results are shown in Supp. Fig~\ref{fig:results}. We also demonstrated that the concept importance learned by our model is better aligned with clinical importance in Supp. Fig~\ref{fig:delta_y_heatmap}. Additionally, we investigated the sensitivity of $\lambda$ on these datasets, as illustrated in Supp. Fig~\ref{fig:lambdas}. 

\textbf{Skin Disease Classification:} We also validated our approach to integrate clinical knowledge into CBM on datasets with skin images$^1$, as shown in Table~\ref{tab:skin_table} and Supp. Fig~\ref{fig:results}. Similar observations were found for the skin datasets, as integrating CBM with clinical knowledge improves the classification performance on out-of-domain datasets. For in-domain validation, while CBM with VGG16 backbones showed improved performance using our approach, CBM with ViT-B/16 and MLP classifiers experienced a slight decrease in performance (<0.3\%). This could be due to the constraint of enforcing the model to learn in a clinical approach, limiting its ability to capture specific patterns within the in-domain dataset. Additionally, there was a drop in performance with CBM using MLP classifiers compared to linear ones, although a similar drop is observed in both our model and baselines. This may be due to overfitting in binary classification.

\textbf{Is knowledge working as expected?} A part of our method randomly removes some input concepts from the classifier, which might act as regularization like Dropout~\cite{srivastava2014dropout}. To ensure that the performance improvement comes from our knowledge integration, we investigate whether random interference, as opposed to systematic integration of knowledge, also improves the model's performance from the baseline or not. The results are shown in Table~\ref{tab:random_alpha}. Remarkably, when clinical knowledge is integrated, the F1-Score for out-of-domain testing achieves a notable improvement compared to the model without the integration of knowledge and with random integration, highlighting the potential benefits of incorporating clinical insights into the classification process.

\begin{table}[tb!]
\centering
\caption{Evaluation on datasets with skin images using F1 score (mean $\pm$ 95\% CI). The results indicate that training the model with clinical knowledge (\textit{Ours}) enhances its classification performance, particularly on the out-of-domain dataset.
}
\label{tab:skin_table}
\begin{tabular}{ccccc}
\toprule
& \multicolumn{2}{c}{\textbf{Fitzpatrick 17k (In-domain)}}                                       & \multicolumn{2}{c}{\textbf{DDI (Out-of-domain)}}                      \\
\cmidrule(lr){2-3} \cmidrule(lr){4-5}
\multirow{-3}{*}{\textbf{\begin{tabular}[c]{@{}c@{}}CBM\\Backbone+Classifier\end{tabular}}} & Baseline & Ours                                                  & Baseline & Ours \\
\midrule
VGG16+Linear                                                                                 & $76.58_{\pm0.80}$        & $\mathbf{77.11}_{\pm0.23}$ & $58.87_{\pm0.39}$        & $\mathbf{60.13}_{\pm0.85}$ \\
VGG16+MLP(20)                                                                                & $74.28_{\pm1.74}$        & $\mathbf{75.64}_{\pm0.91}$                                            & $52.33_{\pm6.01}$        & $\mathbf{54.78}_{\pm7.10}$                                            \\
VGG16+MLP(128)                                                                               & $73.78_{\pm1.49}$        & $\mathbf{75.04}_{\pm0.99}$                                            & $51.60_{\pm7.28}$        & $\mathbf{56.01}_{\pm1.06}$                                            \\
\midrule
ViT-B/16+Linear                                                                              & $76.88_{\pm1.03}$        & $\mathbf{76.97}_{\pm0.86}$                                            & $57.78_{\pm2.54}$        & $\mathbf{59.46}_{\pm1.58}$                                            \\
ViT-B/16+MLP(20)                                                                             & $\mathbf{78.45}_{\pm1.85}$        & $78.17_{\pm0.69}$                                                     & $53.44_{\pm3.44}$        & $\mathbf{57.11}_{\pm4.97}$                                            \\
ViT-B/16+MLP(128)                                                                            & $\mathbf{77.89}_{\pm1.25}$        & $77.70_{\pm1.03}$                                                     & $55.88_{\pm3.52}$        & $\mathbf{59.04}_{\pm3.72}$ \\
\bottomrule
\end{tabular}
\end{table}

\begin{table}[tb!]
\centering
\caption{Comparison of WBC classification F1 score (mean $\pm$ 95\% CI) on out-of-domain testing, with integration (\textit{Ours}), with random integration (\textit{Random}), and without the integration of clinical knowledge (\textit{Baseline}).}
\label{tab:random_alpha}
\begin{tabular}{cccccc}
\toprule
\multicolumn{3}{c}{\textbf{Scirep}} & \multicolumn{3}{c}{\textbf{RaabinWBC}} \\
\cmidrule(lr){1-3} \cmidrule(lr){4-6} 
 Baseline     & Random       & Ours                  & Baseline       & Random         & Ours                   \\
\midrule
 $66.31_{\pm3.73}$ & $67.36_{\pm0.30}$ & $\mathbf{70.50}_{\pm2.94}$ & $54.45_{\pm3.14}$ & $52.68_{\pm3.92}$ & $\mathbf{58.40}_{\pm5.55}$ \\
\bottomrule
\end{tabular}
\end{table}

\section{Conclusion}

Our findings underscore the considerable potential of enhancing the performance of CBMs with clinical knowledge, especially when testing on out-of-domain datasets. The CBM guided by clinical knowledge exhibits better performance when dealing with diverse sources of images, such as WBC images sourced from varied staining conditions and microscopy techniques. Unlike purely data-driven CBMs that rely solely on features extracted from the training set, our clinically guided model incorporates insights accumulated over years of experience. This alignment with clinical knowledge enhances the model’s adaptability across varied image sources, thereby boosting its robustness and predictive accuracy. %

\begin{credits}
\subsubsection{\ackname} 
This research was carried out at the Rapid-Rich Object Search (ROSE) Lab, Nanyang Technological University, Singapore. The research is supported in part by the NTU-PKU Joint Research Institute, a collaboration between Nanyang Technological University and Peking University, sponsored by a donation from the Ng Teng Fong Charitable Foundation. This work is also supported in part by Sysmex Corporation, Japan. We thank Mari Kono and Joao Nunes for their helpful discussions. The computational work for this article was partially performed on the resources of the National Supercomputing Centre (NSCC), Singapore (NSCC project ID: 12003885).

\subsubsection{\discintname}
The authors have no competing interests to declare that are
relevant to the content of this article.
\end{credits}

\bibliographystyle{splncs04}
\bibliography{Paper-1786}{}

\clearpage

\noindent\large\bfseries Supplementary Materials for ``Integrating Clinical Knowledge into Concept
Bottleneck Models''

\begin{table}[h!]
\caption{Classifier layer names and their architectures in PyTorch syntax.}
\centering
\begin{adjustbox}{width=0.95\textwidth}
\begin{tabular}{lc}
\toprule
\textbf{Classifier Name} & \textbf{Architecture} \\
\midrule
Linear & \texttt{nn.Linear(n\_input, n\_classes)} \\
MLP(20) & \texttt{nn.Linear(n\_input, 20) $\rightarrow$ nn.ReLU() $\rightarrow$ nn.Linear(20, n\_classes)} \\
MLP(128) & \texttt{nn.Linear(n\_input, 128) $\rightarrow$ nn.ReLU() $\rightarrow$ nn.Linear(128, n\_classes)} \\
\bottomrule
\end{tabular}
\label{tab:linear}
\end{adjustbox}
\end{table}

\begin{table}[h!]
\centering
\caption{Implementation details. We run the same code three times with different seeds and report 95\% confidence intervals.}
\label{tab:implementations}
\begin{adjustbox}{width=0.8\textwidth}
\begin{tabular}{c|c|c|c|c|c|c|c|c|c}
\toprule
Dataset & \begin{tabular}[c]{@{}c@{}}Backbone\\ +Classifier\end{tabular} & Optimizer & \begin{tabular}[c]{@{}c@{}}Batch\\ size\end{tabular} & Epoch & LR & \begin{tabular}[c]{@{}c@{}}Weight\\ decay\end{tabular} & $\varphi$ & LS & $\lambda$  \\
\midrule
\multirow{6}{*}{WBC} & Vgg16+Linear & \multirow{3}{*}{AdamW} & \multirow{3}{*}{64} & \multirow{3}{*}{30} & \multirow{3}{*}{0.0001} & \multirow{3}{*}{0.01} & \multirow{3}{*}{1} & 0.3 & 1 \\
 & Vgg16+MLP(20) &  &  &  &  &  &  & 0.05 & 1 \\
 & Vgg16+MLP(128) &  &  &  &  &  &  & 0.3 & 1 \\ \cmidrule{2-10} 
 & ViT-B/16+Linear & \multirow{3}{*}{AdamW} & \multirow{3}{*}{64} & \multirow{3}{*}{30} & \multirow{3}{*}{0.0001} & \multirow{3}{*}{0.01} & \multirow{3}{*}{1} & 0.3 & 3 \\
 & ViT-B/16+MLP(20) &  &  &  &  &  &  & 0.1 & 1 \\
 & ViT-B/16+MLP(128) &  &  &  &  &  &  & 0.3 & 1 \\
 \midrule
\multirow{6}{*}{Skin} & Vgg16+Linear & \multirow{3}{*}{AdamW} & \multirow{3}{*}{64} & \multirow{3}{*}{30} & \multirow{3}{*}{0.0001} & \multirow{3}{*}{0.01} & \multirow{3}{*}{1} & 0.1 & 5 \\
 & Vgg16+MLP(20) &  &  &  &  &  &  & 0.1 & 1 \\
 & Vgg16+MLP(128) &  &  &  &  &  &  & 0 & 5 \\ \cmidrule{2-10} 
 & ViT-B/16+Linear & \multirow{3}{*}{AdamW} & \multirow{3}{*}{64} & \multirow{3}{*}{30} & \multirow{3}{*}{0.0001} & \multirow{3}{*}{0.01} & \multirow{3}{*}{1} & 0.5 & 5 \\
 & ViT-B/16+MLP(20) &  &  &  &  &  &  & 0 & 2 \\
 & ViT-B/16+MLP(128) &  &  &  &  &  &  & 0.1 & 1 \\
 \bottomrule
\end{tabular}
\end{adjustbox}
\end{table}

\end{document}